\PassOptionsToPackage{hyphens}{url}
\documentclass{opendatalab}

\usepackage[utf8]{inputenc}
\usepackage[T1]{fontenc}
\usepackage{amsmath}
\usepackage{amssymb}
\usepackage{amsfonts}
\usepackage{tabularx}
\usepackage{nicefrac}
\usepackage{doi}
\usepackage[misc]{ifsym}

\title{From Diagnosis to Correction: Benchmarking and Improving Real-World Table Parsing}

\hypersetup{
    pdftitle={From Diagnosis to Correction: Benchmarking and Improving Real-World Table Parsing},
    pdfauthor={
    	Jutao Xiao, Yuan Qu, Dongsheng Ma, Fan Wu, Tianyao He,
    	Weihong Li, Jie Yang, Yu Qiao, Bin Wang, Conghui He
    }
}

\author[1,2*]{Jutao Xiao}
\author[2*]{Yuan Qu}
\author[3*]{Dongsheng Ma}
\author[2]{Fan Wu}
\author[2]{Tianyao He}
\author[2]{Weihong Li}
\author[2]{Jie Yang}
\author[2 \ \textrm{\Letter}]{Yu Qiao}
\author[2 \ \textrm{\Letter}]{Bin Wang}
\author[2 \ \textrm{\Letter}]{Conghui He}

\affiliation[1]{Zhejiang University}
\affiliation[2]{Shanghai Artificial Intelligence Laboratory}
\affiliation[3]{Peking University}

\contribution[]{xiaojutao@zju.edu.cn, heconghui@pjlab.org.cn}

\abstract{
	Recent document parsers achieve table TEDS scores above 93 on OmniDocBench v1.6, yet community feedback and our audit reveal persistent failures on complex real-world tables. To quantify this gap, we introduce \textbf{TableParseMap}, a diagnostic benchmark of 916 real-world tables organized into five challenging scenarios and nine failure types. The strongest evaluated parser achieves only 85.03 TEDS, showing that aggregate benchmark scores conceal substantial weaknesses. Our analysis attributes these failures to three complementary limitations: large tables exceed the reliable processing scale of a single pass, weak or ambiguous visual cues hinder structure perception, and the reconstructed table may remain visually inconsistent with the image. We therefore propose \textbf{DEC} (Decompose--Enhance--Correct), a visual-consistency-guided agentic framework that improves frozen table parsers without retraining. DEC uses a general VLM as the controller: Decompose partitions large tables along structure-aware boundaries, Enhance exposes weak visual evidence and reparses transformed views, and Correct diagnoses and repairs residual errors. A Visual Consistency Gate (VC-Gate) selectively triggers intervention, while a Visual Consistency Ranker (VC-Ranker) verifies candidate updates and supports rollback without ground-truth HTML at inference time. We further derive a 1,977-table Consensus-Hard Set from 4,556 candidates through offline metrics and cross-model consensus. Across three frozen parsers, DEC improves TEDS by 1.57 points on average; on TableParseMap, gains reach 1.89 points overall, 2.62 on structural errors, and 5.66 on large tables.
}

\metadata[* Equal contribution\quad $\textrm{\Letter}$ Corresponding author]{}

\begin{document}

\maketitle

\section{Introduction}

Recent document parsing models have achieved steadily improving performance on public benchmarks. On OmniDocBench v1.6, PaddleOCR-VL-1.6 and MinerU2.5-Pro obtain overall scores of 96.34 and 95.75, respectively, with table TEDS scores of 94.76 and 93.42 \cite{ouyang2025omnidocbench,wang2026mineru2,zhang2026paddleocr}. However, these strong benchmark results do not fully match users' experience in real-world deployments. We collect table-recognition-related issues from the MinerU community and find frequent reports of large tables, borderless layouts, missing content, duplicated outputs, row--column misalignment, and incorrect merged cells.
\begin{figure}[t]
	\centering
	\includegraphics[
	width=\columnwidth,
	]{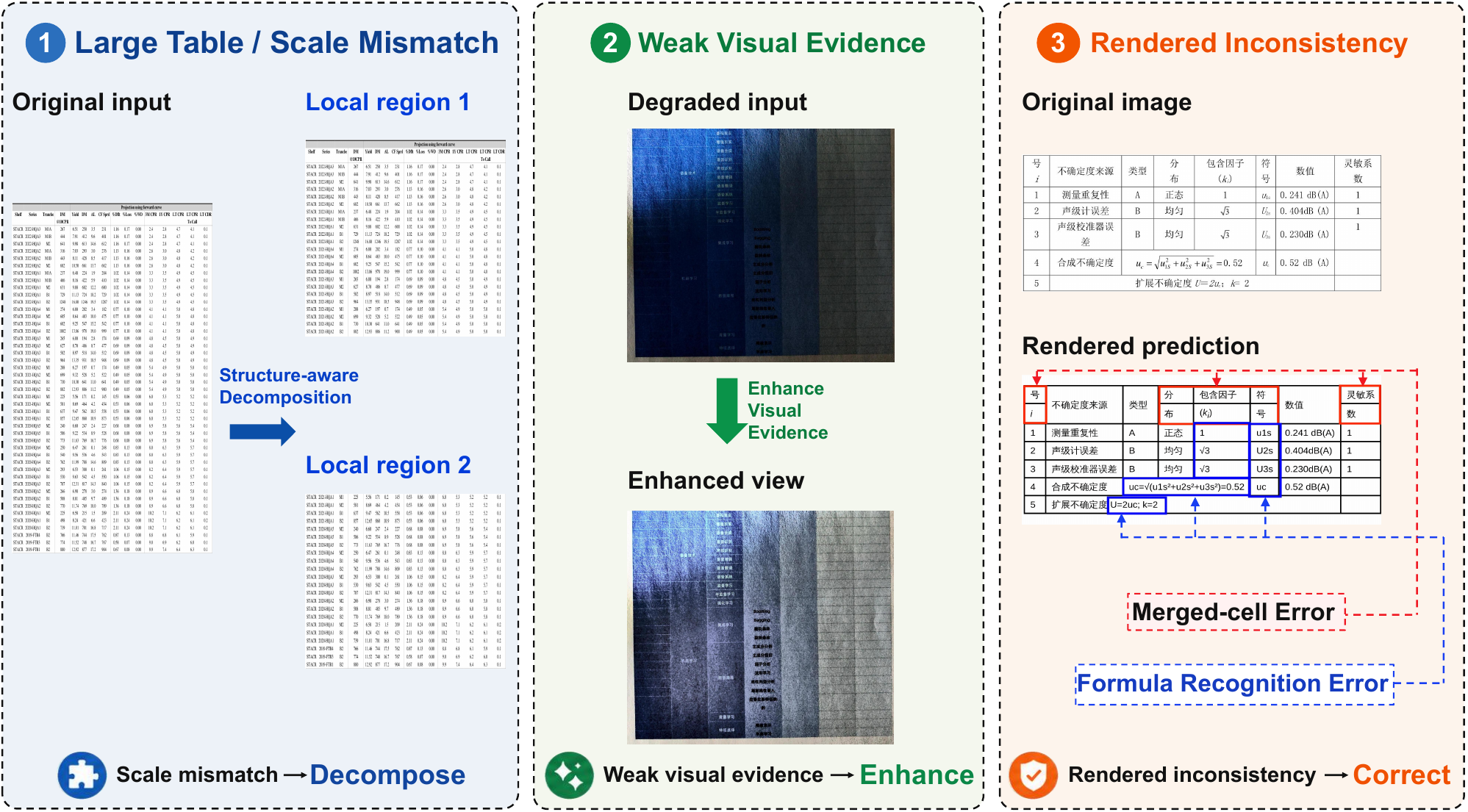}
	\caption{Representative DEC interventions for scale mismatch, weak visual evidence, and rendered inconsistency.}
	\label{fig:dec-intervention}
\end{figure}

To quantify and explain this gap, we construct TableParseMap, a fine-grained diagnostic benchmark containing 916 complex table images from real-world applications. We systematically review the outputs of MinerU2.5-Pro, PaddleOCR-VL-1.6, GLM-OCR \cite{duan2026glm}, and Qwen3.5-397B-A17B \cite{team2026qwen3}, and organize recurring problems into five challenging scenarios and nine failure types. The five scenarios characterize where parsers are likely to struggle, while the nine failure types describe how they fail through recurring structural and textual errors such as omission, misalignment, duplicated output, and incorrect merged cells. TableParseMap thus exposes model behaviors hidden by aggregate scores from two complementary perspectives: input scenarios and output failures. On this benchmark, MinerU2.5-Pro, the strongest evaluated model, achieves only 85.03 TEDS and falls below 80 on the Implicit Semantic scenario.

The failures revealed by TableParseMap point to three complementary limitations. First, finite input resolution, visual-token capacity, and output length make large tables difficult to process completely in a single pass. Second, weak or ambiguous visual cues may prevent reliable structure recovery even when the full table is visible. Third, the reconstructed table may remain visually inconsistent with the image. Addressing these long-tail failures through continual data collection and retraining requires additional annotation, computation, and access to model parameters, making it unsuitable for fixed deployments or closed-source systems. Existing split-and-merge methods are typically embedded within dedicated table recognition networks \cite{zhang2022split,guo2022trust,qin2024semv3}, while recent agentic methods mainly diagnose and revise outputs after initial parsing \cite{yu2026parsefixer,wen2026ocr}. The former cannot be readily applied to frozen parsers, whereas the latter cannot recover content or visual evidence missed during the initial parsing process.

Based on these observations, we propose DEC (Decompose--Enhance--Correct), a visual-consistency-guided agentic framework that requires no retraining of the base parser, with representative interventions illustrated in Fig.~\ref{fig:dec-intervention}. \textbf{Decompose} selects structure-aware split points and partitions large tables into local blocks that the parser can process reliably. \textbf{Enhance} uses image-based tools to expose weak structural cues and re-invokes the frozen parser. \textbf{Correct} compares the table image with the current rendered prediction to diagnose and repair residual inconsistencies. DEC further introduces a Visual Consistency Gate (VC-Gate) to determine whether intervention is needed and a Visual Consistency Ranker (VC-Ranker) to verify candidate updates and support rollback. Together, they enable selective and reliable intervention without ground-truth HTML at inference time.

To evaluate DEC on samples that challenge multiple parsers, we design an automated screening protocol inspired by the Cross-Model Consistency Verification strategy in MinerU2.5-Pro. The protocol combines offline metrics and cross-model consensus to identify samples that remain difficult across heterogeneous parsers. We apply it to 4,556 candidate samples drawn from public benchmarks, real-world data, and TableParseMap, yielding a Consensus-Hard Set of 1,977 tables. Five representative baselines achieve only 66.10--70.92 TEDS on this set, indicating that it remains challenging across different parsers. Because screening is driven by unified metrics and model outputs rather than a fixed manually selected list, the protocol can be rerun as parsers improve to refresh the cross-model difficulty boundary.

Our experiments cover three frozen parsers and two controller scales. Across all evaluation settings, DEC improves TEDS by 1.57 points on average. On TableParseMap, it improves the overall TEDS of the three base parsers by an average of 1.89 points, with average gains of 2.62 points on structural errors and 5.66 points on Large Scale Tables. Fine-grained analyses and ablation studies further validate the contributions of the three stages, the visual tools, and the runtime decision modules.

Our main contributions are summarized as follows:

\begin{itemize}
	\item \textbf{Benchmark.} We introduce TableParseMap, a diagnostic benchmark of 916 real-world complex tables with five challenging scenarios and nine failure types for fine-grained parser analysis.

	\item \textbf{Framework.} We propose DEC, a visual-consistency-guided agentic framework for progressive inference-time intervention over frozen table parsers, with VC-Gate and VC-Ranker enabling selective routing, verification, and rollback without ground-truth HTML.
	
	\item \textbf{Evaluation.} We derive a 1,977-table Consensus-Hard Set through automated cross-model screening and evaluate DEC across three frozen parsers and two controller scales, achieving an average TEDS gain of 1.57 points.
	
\end{itemize}

\begin{figure*}[t]
	\centering
	\includegraphics[width=\textwidth]{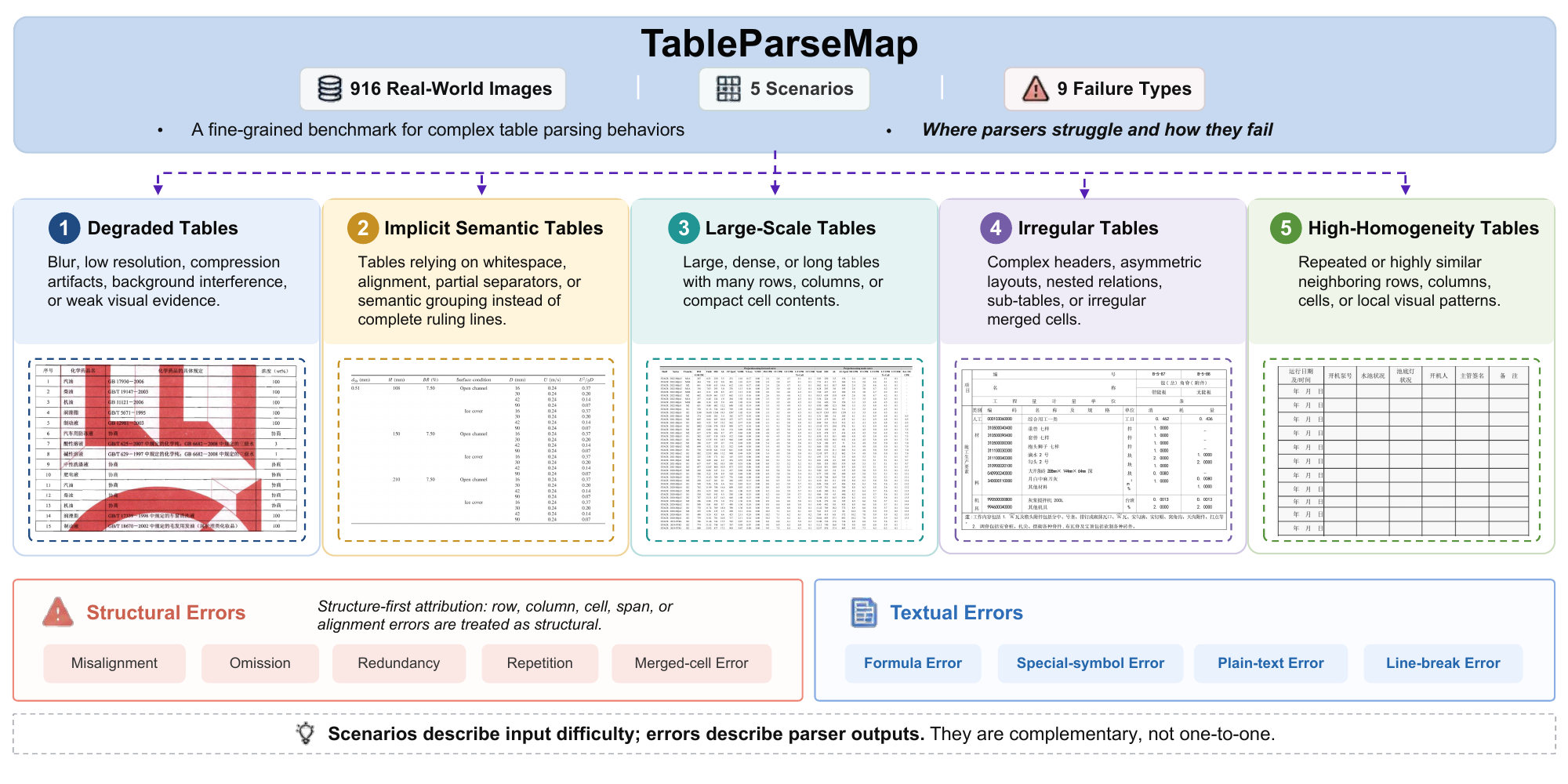}
	\caption{Overview of TableParseMap. The benchmark contains 916
		real-world table images and characterizes parser behavior through five
		challenging input scenarios and nine failure types. The scenario
		taxonomy describes where parsers struggle, while the structure-first
		error taxonomy describes how they fail.}
	\label{fig:tableparsemap}
\end{figure*}
\section{Related Work}

\subsection{Table Parsing Models}

Table parsing has evolved from handcrafted line and coordinate rules to data-driven detection and structure recognition \cite{schreiber2017deepdesrt,zheng2020global}. Later methods model relations, grids, logical coordinates, object queries, or HTML sequences to recover two-dimensional structure \cite{zhong2020image,zhang2022split,nassar2022tableformer,raja2022visual,lin2022tsrformer,wang2023robust,xing2023lore,huang2023improving}. Recent table and document VLMs further unify table recognition with broader visual table understanding or full-page parsing \cite{zhao2024tabpedia,liu2024grab,peng2024unitable,zhang2024unitabnet,niu2025mineru2,cui2025paddleocr}, although general VLMs still exhibit limitations in spatial and formatting understanding \cite{xia2024vision}. Instruction-guided methods further improve difficult table structures through dedicated training \cite{chen2026instructtable}. Most improvements, however, require retraining and cannot directly extend deployed or closed-source parsers. DEC instead keeps the parser frozen and improves difficult cases through inference-time decomposition, enhancement, and correction.

\subsection{Table Benchmarks}

ICDAR 2013 and cTDaR standardized early table detection and structure recognition, while PubTabNet and PubTables-1M expanded image-to-HTML evaluation and annotation scale \cite{gobel2013icdar,gao2019icdar,zhong2020image,smock2022pubtables}. Recent document and OCR benchmarks broaden domain and acquisition coverage \cite{ouyang2025omnidocbench,yang2025cc,fu2026ocrbench,li2026towards,zhou2026real5}, but primarily report aggregate scores or coarse groups. TableParseMap complements them by jointly characterizing challenging input scenarios and resulting parser failures, while the Consensus-Hard Set concentrates cross-model difficult cases for targeted evaluation.

\subsection{Agentic Parsing and Runtime Verification}

Test-time refinement improves outputs through iterative critique and revision without retraining \cite{madaan2023self}. Text-based post-OCR correction further repairs recognition and reading-order errors using dedicated correction models \cite{shim2025revise}. OCR-Agent and ParseFixer apply iterative refinement to OCR and document parsing, with ParseFixer combining a frozen parser with selective correction and rollback \cite{wen2026ocr,yu2026parsefixer}. Multi-model agreement has also been explored as a training-free signal for sample-level OCR reliability, output selection, and adaptive routing \cite{zhang2025consensus}. Related work also uses render-based or visual-equivalence signals for table evaluation and refinement \cite{zhang2025monkeyocr,zhang2025docr,liu2026visual}. Existing correction and agentic methods remain largely post-hoc, while visual verification is often used only as a reward or final evaluator. DEC couples them throughout inference, using visual consistency to control decomposition, evidence enhancement, correction, and update acceptance.
\section{TableParseMap}

We introduce TableParseMap, a diagnostic benchmark designed to quantify
where table parsers struggle and how they fail on complex real-world
tables. Unlike benchmarks that summarize performance with a single
aggregate score, TableParseMap organizes parser behavior from two
complementary views: challenging input scenarios and observed output
failures.

\subsection{Benchmark Construction}

\paragraph{Model Audit and Sample Selection.}
We construct TableParseMap by systematically auditing real parser failures. We run MinerU2.5-Pro, PaddleOCR-VL-1.6, GLM-OCR, and Qwen3.5-397B-A17B on real-world table data and manually compare their outputs with the original images. Consolidating recurring failures across the four models yields nine output error types, which we associate with five high-risk input scenarios. We then sample tables targeting these scenarios and assign each image one canonical tag through expert review, producing 916 annotated images. Because this process intentionally targets recurrent failures, TableParseMap supports diagnostic analysis rather than accuracy estimation under a natural table distribution.

\subsection{Taxonomy and Annotation Protocol}

\paragraph{Dual-View Taxonomy.}
Each image receives exactly one of 14 canonical tags, grouped into text
errors, structural errors, and challenging input scenarios. The 331
text-error samples cover character recognition, line breaks, formulas,
and special symbols. The 307 structural-error samples cover omission,
misalignment, redundancy, duplicated output, and merged-cell errors. The
remaining 278 images form five scenario categories: implicit semantic
tables, degraded visual evidence, highly similar patterns, irregular or
rare structures, and large tables. Scenario tags describe input
conditions under which parsers struggle, whereas error tags describe
failures observed in parser outputs. Figure~\ref{fig:tableparsemap} summarizes the five challenging input scenarios, while representative examples of all nine parser failure types are provided in Appendix~\ref{app:tableparsemap_cases}.

\paragraph{Structure-First Annotation.}
When multiple phenomena appear in the same sample, annotators assign the
tag corresponding to the dominant failure or input condition. We follow
a structure-first rule to avoid assigning multiple labels to a single
underlying error. For example, if a missing cell also causes its text to
disappear, the sample is labeled as a structural omission because the
text loss is a consequence of the structural error. A text-error label is
used only when the table structure is correct and the cell content itself
is misrecognized.

\begin{figure*}[t]
	\centering
	\includegraphics[width=\textwidth]{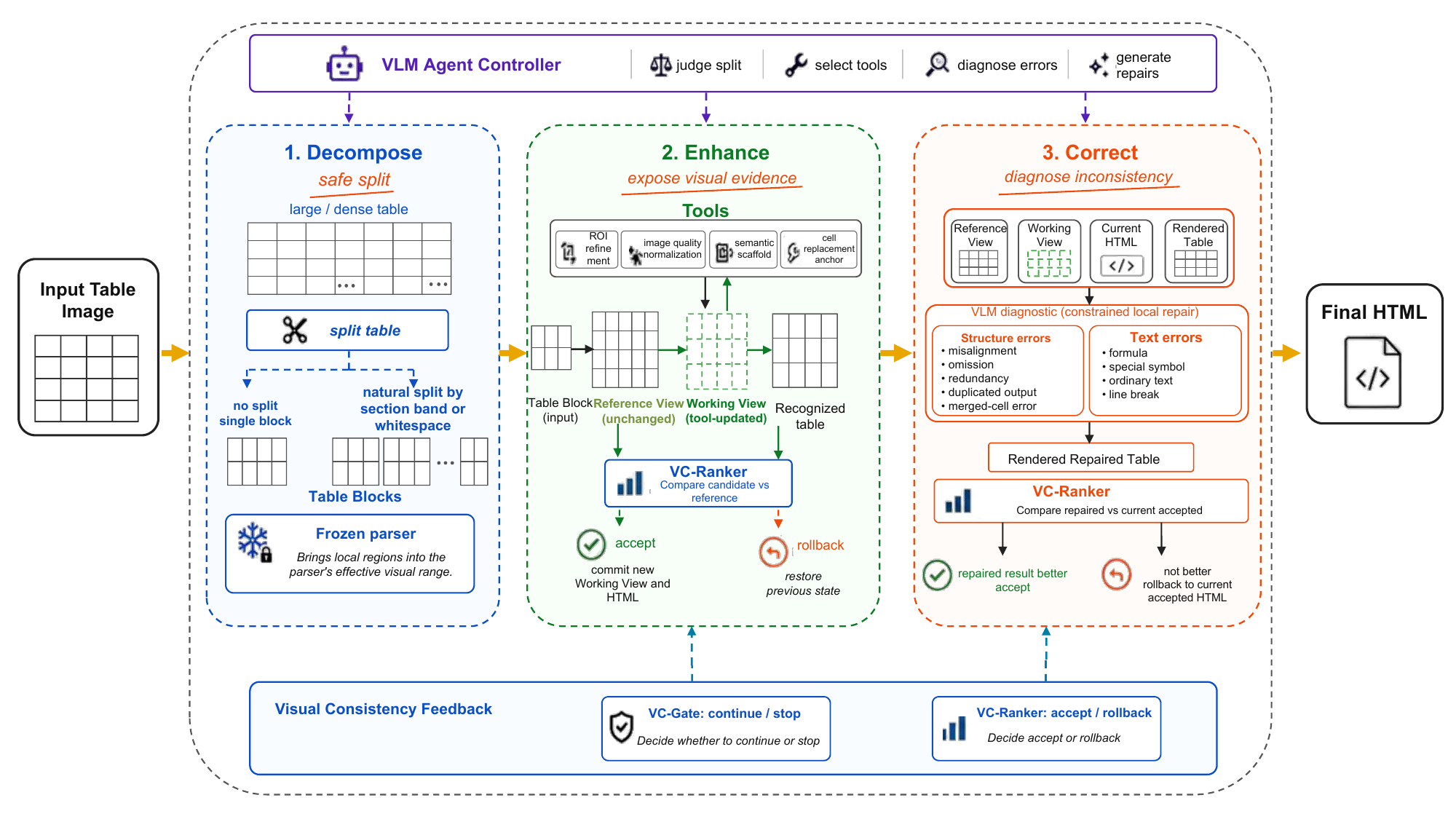}
	\caption{Overview of DEC. A general VLM controls Decompose, Enhance, and Correct over a frozen table parser, while VC-Gate and VC-Ranker govern stage triggering, candidate acceptance, and fallback.}
	\label{fig:framework}
\end{figure*}

\section{Method}
\subsection{Overview}

Given a table image $I$, a frozen base parser $R_{\theta}$ first produces an initial structured representation, such as HTML, $Y_{\mathrm{base}}=R_{\theta}(I)$. Building on the three recurring limitations revealed by TableParseMap---scale mismatch, weak or ambiguous visual evidence, and residual inconsistency between the reconstructed table and the input image---we propose Decompose--Enhance--Correct (DEC), a progressive inference-time framework over the frozen parser.

DEC addresses these limitations through three stages. \textbf{Decompose} partitions large tables into locally manageable regions; \textbf{Enhance} exposes existing but difficult-to-recognize visual evidence through image-based tools; and \textbf{Correct} diagnoses and repairs residual inconsistencies by jointly inspecting the reference image, the current structured prediction, and its rendering. DEC uses the frozen parser $R_{\theta}$ as the recognition executor and a general vision-language model $A_{\phi}$ as the controller. VC-Gate $G_{\eta}$ determines whether further intervention is required, whereas VC-Ranker $S_{\omega}$ evaluates whether a tentative update improves visual consistency and governs candidate acceptance and fallback.

\subsection{Visual-Consistency Decision Models}

At inference time, ground-truth HTML is unavailable. DEC therefore controls stage triggering and state updates through the visual consistency between the original table image $I$ and the deterministic rendering $\rho(Y)$ of a parsing result $Y$. We decouple this ground-truth-free quality assessment into two complementary decisions: whether the current result is acceptable and whether a tentative result is better than the current one.

VC-Gate $G_{\eta}$ predicts the acceptability of a result, while VC-Ranker $S_{\omega}$ assigns a relative visual-consistency score under the same reference image:
\begin{equation}
	g_{\eta}(Y\mid I)=G_{\eta}\bigl(I,\rho(Y)\bigr),s_{\omega}(Y\mid I)=S_{\omega}\bigl(I,\rho(Y)\bigr),
\end{equation}
where $g_{\eta}(Y\mid I)\in\{0,1\}$. VC-Gate determines whether the base result should enter DEC and whether the output of Enhance still requires Correct.
For a current result $Y$ and a tentative result $\widetilde{Y}$, DEC defines the acceptance decision as
\begin{equation}
	\label{eq:accept}
	\operatorname{Accept}(\widetilde{Y},Y\mid I)
	=
	\mathbb{I}\left[
	s_{\omega}(\widetilde{Y}\mid I)
	>
	s_{\omega}(Y\mid I)+\delta
	\right]
\end{equation}
This rule governs candidate acceptance in Enhance and Correct and is also used for post-merge verification.

\subsection{Decompose--Enhance--Correct}

\subsubsection{Decompose}

Decompose handles tables whose scale or density exceeds the effective visual range of the base parser. Rather than uniformly downscaling the input or applying arbitrary crops, the agent invokes \texttt{split\_table} to identify structurally safe boundaries from whitespace and subtable cues while avoiding cuts through cells or merged regions. This process produces $K$ reference image blocks:
\begin{equation}
	\label{eq:decompose}
	\left\{I_k^{\mathrm{ref}}\right\}_{k=1}^{K}
	=
	\mathcal{D}\left(I;A_{\phi}\right)
\end{equation}
Tables that do not require decomposition are treated as $K=1$ with $I_1^{\mathrm{ref}}=I$. Each block is subsequently processed by Enhance and Correct, and the block-level results are merged only when $K>1$.

\subsubsection{Enhance}

Enhance exposes visual evidence that is present in the image but difficult for the base parser to recognize. For each block, DEC maintains an immutable \emph{Reference View} $I_k^{\mathrm{ref}}$ for visual-consistency evaluation and an editable \emph{Working View} $I_k^{(r)}$ for tool-based enhancement, initialized as $I_k^{(0)}=I_k^{\mathrm{ref}}$. The corresponding initial parsing result is defined as
\begin{equation}
	\label{eq:enhance_init}
	Y_{k,\mathrm{E}}^{(0)}
	=
	\begin{cases}
		Y_{\mathrm{base}}, & K=1,\\
		R_{\theta}\bigl(I_k^{\mathrm{ref}}\bigr), & K>1
	\end{cases}
\end{equation}
The enhancement tools perform table-region refinement, image-quality normalization, sparse structural scaffolding, and removable cell anchoring. These transformations expose existing visual evidence without adding target text or directly modifying the candidate HTML.

At round $r$, the agent selects an enhancement action $a_{k,r}$ based on the Reference View, the current Working View, and previous tool feedback. The selected transformation generates a tentative Working View, which is reparsed by the frozen parser:
\begin{equation}
	\label{eq:enhance}
	\widetilde{I}_k^{(r+1)}
	=
	T_{a_{k,r}}\bigl(I_k^{(r)}\bigr),
	\qquad
	\widetilde{Y}_{k,\mathrm{E}}^{(r+1)}
	=
	R_{\theta}\bigl(\widetilde{I}_k^{(r+1)}\bigr)
\end{equation}
Using $I_k^{\mathrm{ref}}$ as the evaluation reference, DEC evaluates the tentative image--HTML state according to Eq.~\eqref{eq:accept}. If accepted, it commits the tentative state; otherwise, it rolls back to the previous state. Enhance terminates when no further tool action is proposed or the maximum number of rounds is reached. The retained Working View and parsing result are denoted by $I_k^{\mathrm{work}}$ and $Y_k^{\mathrm{enh}}$, respectively.

\subsubsection{Correct}

After Enhance, VC-Gate evaluates $Y_k^{\mathrm{enh}}$. If the result is accepted, DEC skips correction and sets $Y_k^{\mathrm{final}}=Y_k^{\mathrm{enh}}$; otherwise, it initializes the correction trajectory as $Y_{k,\mathrm{C}}^{(0)}=Y_k^{\mathrm{enh}}$. At each round $t$, the agent jointly examines the Reference View, Working View, current HTML, and its rendering. It first diagnoses image--HTML inconsistencies and then generates a corrected candidate conditioned on the diagnosis:
\begin{equation}
	\label{eq:correct}
	\begin{aligned}
		D_k^{(t)}
		&=
		A_{\phi}^{\mathrm{diag}}\left(
		I_k^{\mathrm{ref}},
		I_k^{\mathrm{work}},
		Y_{k,\mathrm{C}}^{(t)},
		\rho\bigl(Y_{k,\mathrm{C}}^{(t)}\bigr)
		\right),\\
		\widetilde{Y}_{k,\mathrm{C}}^{(t+1)}
		&=
		A_{\phi}^{\mathrm{repair}}\left(
		I_k^{\mathrm{ref}},
		I_k^{\mathrm{work}},
		Y_{k,\mathrm{C}}^{(t)},
		D_k^{(t)}
		\right)
	\end{aligned}
\end{equation}
Here, $A_{\phi}^{\mathrm{diag}}$ and $A_{\phi}^{\mathrm{repair}}$ are two prompt-conditioned calls to the same VLM rather than separate models. Diagnosis is guided by the predefined failure taxonomy, while repair generates a revised complete HTML grounded in the available visual evidence. No sample-level scene or error labels from TableParseMap are used during inference, and the agent may abstain when the available visual evidence is insufficient.

If $\operatorname{Accept}(\widetilde{Y}_{k,\mathrm{C}}^{(t+1)},Y_{k,\mathrm{C}}^{(t)}\mid I_k^{\mathrm{ref}})=1$, DEC commits the corrected candidate as $Y_{k,\mathrm{C}}^{(t+1)}$; otherwise, it retains $Y_{k,\mathrm{C}}^{(t)}$ and terminates the current correction trajectory. Correct also stops when no further error is diagnosed or the maximum number of rounds is reached. The retained correction result is denoted by $Y_k^{\mathrm{final}}$.

\subsection{Merge and Final Verification}

When $K=1$, DEC directly returns the block-level result $Y_1^{\mathrm{final}}$. When $K>1$, the independently processed HTML results are temporarily converted to OTSL\cite{lysak2023optimized}, aligned and concatenated according to the original spatial arrangement $\mathcal{P}$ of the image blocks, and then converted back to HTML:
\begin{equation}
	\label{eq:merge}
	Y_{\mathrm{merge}}
	=
	\psi_{\mathrm{O\rightarrow H}}\!\Bigl(
	\mathcal{M}_{\mathrm{OTSL}}\!\bigl(
	\{\psi_{\mathrm{H\rightarrow O}}(Y_k^{\mathrm{final}})\}_{k=1}^{K};
	\mathcal{P}
	\bigr)
	\Bigr)
\end{equation}

Although each block has been verified independently, merging may introduce global inconsistencies such as duplicated boundary rows, missing content, or cross-block misalignment. DEC therefore performs post-merge verification only when Decompose is triggered. Using the complete image $I$ as the reference, VC-Ranker compares $Y_{\mathrm{merge}}$ with the initial full-table result $Y_{\mathrm{base}}$, and the final output is
\begin{equation}
	\label{eq:final_output}
	Y^{*}
	=
	\begin{cases}
		Y_{\mathrm{merge}},
		& \operatorname{Accept}\!\left(
		Y_{\mathrm{merge}},Y_{\mathrm{base}}\mid I
		\right)=1,\\
		Y_{\mathrm{base}},
		& \text{otherwise}
	\end{cases}
\end{equation}
This verification requires one additional rendering and one VC-Ranker inference, without rerunning the base parser.

\begin{table*}[t]
	\centering
	\small
	\setlength{\tabcolsep}{0.6mm}
	\renewcommand{\arraystretch}{1.06}
	
	\resizebox{\textwidth}{!}{%
	\begin{tabular}{@{}ll*{7}{c}@{}}
		\toprule
		Parser
		& DEC Controller
		& Overall
		& CCOCR
		& In-house
		& OCRBenchv2
		& Real5
		& Wild
		& \shortstack{TableParse\\Map} \\
		\midrule
		
		Qwen3.5-397B
		& --
		& 66.52/76.68
		& 76.20/80.81
		& 68.84/79.89
		& 73.22/78.63
		& 54.89/71.02
		& 53.24/67.57
		& 74.86/80.93 \\
		
		Qwen3.5-35B
		& --
		& 66.10/76.45
		& 76.98/82.54
		& 66.78/78.47
		& 70.20/76.25
		& 54.00/70.19
		& 51.82/66.52
		& 76.72/82.55 \\
		\midrule
		
		\multirow{3}{*}{MinerU2.5-Pro}
		& --
		& 70.92/81.11
		& 75.62/81.54
		& 75.84/85.73
		& 84.49/90.25
		& 56.74/73.12
		& 53.68/68.21
		& 79.38/85.39 \\
		
		& Qwen3.5-397B
		& 72.18/82.19
		& 77.52/83.08
		& 76.98/86.67
		& 85.69/91.39
		& 57.57/73.74
		& 55.76/69.79
		& 80.91/86.79 \\
		
		& Qwen3.5-35B
		& 72.14/82.11
		& 76.89/83.01
		& 76.96/86.59
		& 85.36/91.00
		& 57.86/73.84
		& 55.21/69.41
		& 80.85/86.72 \\
		\midrule
		
		\multirow{3}{*}{PaddleOCR-VL-1.6}
		& --
		& 67.50/77.90
		& 66.57/75.00
		& 69.25/80.54
		& 75.72/82.49
		& 57.61/72.42
		& 54.40/68.18
		& 75.74/82.57 \\
		
		& Qwen3.5-397B
		& 69.41/79.58
		& 69.77/77.32
		& 70.80/82.04
		& 79.03/85.30
		& 58.32/73.22
		& 56.35/69.14
		& 78.17/84.74 \\
		
		& Qwen3.5-35B
		& 68.82/78.96
		& 69.90/77.46
		& 70.80/82.02
		& 77.21/83.53
		& 58.29/72.97
		& 56.18/68.89
		& 77.05/83.77 \\
		\midrule
		
		\multirow{3}{*}{GLM-OCR}
		& --
		& 69.13/79.80
		& 77.20/83.92
		& 72.25/83.33
		& 79.52/86.57
		& 57.74/73.36
		& 55.40/69.35
		& 75.66/82.84 \\
		
		& Qwen3.5-397B
		& 70.68/81.01
		& 78.78/84.98
		& 73.87/84.54
		& 82.39/88.82
		& 58.19/73.65
		& 56.22/69.69
		& 77.86/84.75 \\
		
		& Qwen3.5-35B
		& 70.13/80.57
		& 79.38/84.96
		& 72.98/83.86
		& 81.10/87.63
		& 58.44/74.02
		& 56.17/69.85
		& 76.68/83.73 \\
		\bottomrule
	\end{tabular}%
	}
	
	\caption{Results on the 1,977-table Consensus-Hard Set. Cells report TEDS/TEDS-S, and Overall is sample-weighted across the six sources. Qwen3.5-397B/35B denote Qwen3.5-397B-A17B/35B-A3B; TableParseMap contains its 602-sample subset.}
	\label{tab:main}
\end{table*}

\begin{table*}[t]
	\centering
	\small
	\setlength{\tabcolsep}{0.35mm}
	\renewcommand{\arraystretch}{1.04}
	
	\resizebox{\textwidth}{!}{%
	\begin{tabular}{@{}l*{8}{c}@{}}
		\toprule
		Parser
		& \shortstack{Overall\\($N=916$)}
		& \shortstack{Text\\Errors\\($N=331$)}
		& \shortstack{Structural\\Errors\\($N=307$)}
		& \shortstack{Implicit\\Semantic\\($N=52$)}
		& \shortstack{Degraded\\Tables\\($N=66$)}
		& \shortstack{High\\Homogeneity\\($N=57$)}
		& \shortstack{Irregular\\Tables\\($N=52$)}
		& \shortstack{Large-Scale\\Tables\\($N=51$)} \\
		\midrule
		
		MinerU2.5-Pro
		& \underline{85.03}/\underline{89.73}
		& \textbf{86.13}/\textbf{93.04}
		& \underline{84.93}/88.22
		& 78.12/84.35
		& 84.92/89.03
		& 86.57/88.42
		& \underline{82.26}/\underline{85.69}
		& 86.67/89.32 \\
		
		PaddleOCR-VL-1.6
		& 81.86/87.22
		& 83.09/89.93
		& 82.01/86.47
		& 71.68/79.42
		& 84.79/89.07
		& 81.86/83.18
		& 78.90/82.71
		& 82.56/88.84 \\
		
		GLM-OCR
		& 79.50/84.88
		& 81.61/89.03
		& 80.51/84.31
		& 72.48/80.01
		& 79.99/85.63
		& 71.08/72.95
		& 76.82/81.00
		& 78.38/82.68 \\
		
		Qwen3.5-397B
		& 80.78/85.72
		& 79.73/87.47
		& 73.39/77.31
		& \textbf{79.18}/\textbf{84.93}
		& 83.74/87.38
		& 76.14/77.58
		& 73.88/77.05
		& \underline{92.26}/\underline{94.30} \\
		
		Qwen3.5-35B
		& 81.21/86.11
		& 78.98/85.90
		& 83.45/87.33
		& 76.00/82.60
		& 83.20/86.36
		& 81.29/82.39
		& 77.07/81.40
		& 89.03/92.29 \\
		DeepSeek-OCR-2
		& 54.36/62.56
		& 60.83/70.25
		& 56.09/63.46
		& 56.51/66.55
		& 51.90/60.84
		& 36.52/40.68
		& 44.60/53.47
		& 32.89/39.18 \\
		
		Unlimited-OCR
		& 64.15/70.17
		& 62.86/70.20
		& 66.97/71.93
		& 64.90/72.84
		& 67.99/73.06
		& 60.22/63.46
		& 63.58/67.13
		& 54.78/63.59 \\
		
		Youtu-Parsing
		& 77.71/84.04
		& 76.86/85.51
		& 78.90/83.73
		& 72.95/80.50
		& 76.63/83.40
		& \underline{86.66}/\underline{89.31}
		& 78.25/83.41
		& 71.87/75.61 \\
		\midrule
		
		MinerU2.5-Pro $+$ DEC
		& \textbf{86.19}/\textbf{90.79}
		& \underline{85.89}/\underline{92.86}
		& \textbf{86.70}/\textbf{89.98}
		& \underline{78.35}/\underline{84.61}
		& \underline{86.49}/\underline{90.35}
		& \textbf{89.32}/\textbf{90.38}
		& \textbf{82.67}/\textbf{86.09}
		& \textbf{92.69}/\textbf{94.37} \\
		
		PaddleOCR-VL-1.6 $+$ DEC
		& 83.76/88.89
		& 83.55/90.45
		& 84.65/\underline{88.77}
		& 73.58/80.87
		& \textbf{87.61}/\textbf{91.27}
		& 84.53/85.84
		& 80.51/84.26
		& 87.58/92.65 \\
		
		GLM-OCR $+$ DEC
		& 82.11/87.47
		& 82.55/90.18
		& 83.96/87.68
		& 74.60/82.16
		& 80.17/85.24
		& 77.92/79.90
		& 80.82/84.29
		& 84.32/88.49 \\
		
		\bottomrule
	\end{tabular}%
	}
	
	\caption{Fine-grained TEDS/TEDS-S results on TableParseMap. Text and Structural are aggregate error groups; the remaining columns are five challenging scenarios. $+$DEC uses Qwen3.5-397B with VC-Gate. Best and second-best values are bold and underlined.}
	
	\label{tab:tableparsemap-finegrained}
\end{table*}

\section{Experiments}
\subsection{Experimental Setup}

\paragraph{Evaluation protocol.}
We evaluate DEC under three complementary settings. Main results use the 1,977-table Consensus-Hard Set, fine-grained analysis uses the complete 916-table TableParseMap, and routing efficiency is measured on the complete six-source pool of 4,556 tables.

\paragraph{Consensus-Hard Set.}
Starting from 4,556 candidates drawn from CCOCR, In-house data, OCRBenchv2, Real5-OmniDocBench, Wild-OmniDocBench, and TableParseMap, we run MinerU2.5-Pro, PaddleOCR-VL-1.6, GLM-OCR, and Qwen3.5-397B-A17B. Selection is independent of VC-Gate: a sample is retained when at least two parsers obtain TEDS below $0.90$, followed by stratified sampling across the six sources to form the final 1,977-table set. This consensus rule avoids tailoring evaluation to isolated failures of a single parser. Because the resulting set targets difficult cases rather than the original data distribution, we also report results on the complete public benchmarks.

\paragraph{Models and metrics.}
We evaluate MinerU2.5-Pro, PaddleOCR-VL-1.6, and GLM-OCR as frozen base parsers, each paired with Qwen3.5-397B-A17B or Qwen3.5-35B-A3B as the DEC controller. For direct comparison on TableParseMap, we additionally include both Qwen models, DeepSeek-OCR-2 \cite{wei2026deepseek}, Unlimited-OCR \cite{yin2026unlimited}, and YoutuParsing \cite{yin2026youtu}. Unless otherwise specified, DEC uses Qwen3.5-397B-A17B as the controller, and mechanism ablations use PaddleOCR-VL-1.6 as the base parser. We report TEDS and TEDS-S on a 0--100 scale using shared annotations, HTML normalization, and metric implementation; Overall on Consensus-Hard is sample-weighted.

\paragraph{Inference configuration.}
Both controllers use deterministic decoding with reasoning disabled. Correct runs for at most three rounds, and candidate updates are accepted using a VC-Ranker margin of $\delta=0.05$. All routing and acceptance thresholds are fixed before testing. VC-Gate and VC-Ranker are trained on data disjoint from all evaluation sets, and neither model accesses ground-truth HTML or offline TEDS at inference time.

\noindent
Further details on source preprocessing, Consensus-Hard construction, VC-Gate and VC-Ranker training, metric aggregation, and efficiency accounting are provided in Appendix~\ref{app:training_details}.

\begin{table*}[t]
	\centering
	\small
	\renewcommand{\arraystretch}{1.05}
	
	\begin{minipage}[t]{0.40\textwidth}
		\centering
		\textbf{(a) Stage-wise Progression}\\[2pt]
		\setlength{\tabcolsep}{1.3pt}
		\resizebox{\linewidth}{!}{%
		\begin{tabular}{@{}lccc@{}}
			\toprule
			Setting & TEDS & TEDS-S & \shortstack{Incremental Gain\\TEDS / TEDS-S} \\
			\midrule
			Parser Baseline & 67.503 & 77.896 & -- \\
			$+$ Decompose & 67.791 & 78.128 & $+0.288$ / $+0.232$ \\
			$+$ Enhance & 68.323 & 78.548 & $+0.532$ / $+0.420$ \\
			Full DEC & \textbf{69.410} & \textbf{79.576} & $\mathbf{+1.087}$ / $\mathbf{+1.028}$ \\
			\bottomrule
		\end{tabular}%
		}
	\end{minipage}
	\hfill
	\begin{minipage}[t]{0.58\textwidth}
		\centering
		\textbf{(b) Update-Selection Strategy}\\[2pt]
		\setlength{\tabcolsep}{1.5pt}
		\begin{tabular}{@{}lccc@{}}
			\toprule
			Strategy & \shortstack{After Enhance\\TEDS / TEDS-S} & \shortstack{Final\\TEDS / TEDS-S} & Reg. Rate \\
			\midrule
			Accept All & 66.816 / 77.497 & 68.549 / 79.341 & 29.34\% \\
			Agent Self-Judge & 66.938 / 77.562 & 68.897 / 79.257 & 13.45\% \\
			\textbf{VC-Ranker} & \textbf{68.323 / 78.548} & \textbf{69.410 / 79.576} & \textbf{12.65\%} \\
			\bottomrule
		\end{tabular}
	\end{minipage}
	
\caption{Mechanism ablations on the Consensus-Hard Set. (a) Stage-wise progression. (b) Update-selection strategies; Reg. Rate is the proportion of samples falling below the shared parser baseline.}
	\label{tab:mechanism-ablation}
\end{table*}
\begin{table*}[t]
	\centering
	\small
	\setlength{\tabcolsep}{1.2mm}
	\begin{tabular}{@{}lcccc@{}}
		\toprule
		Setting
		& TEDS / TEDS-S $\uparrow$
		& \shortstack{DEC Trigger\\Rate}
		& \shortstack{Extra Parser Calls\\per Input $\downarrow$}
		& \shortstack{Controller Calls\\per Input $\downarrow$} \\
		\midrule
		PaddleOCR-VL-1.6 (No DEC)
		& 82.988 / 88.420
		& 0\%
		& 0.000
		& 0.000 \\
		Always-on DEC
		& \textbf{84.449 / 89.641}
		& 100\%
		& 0.924
		& 4.185 \\
		VC-Gate $+$ DEC
		& 84.413 / 89.614
		& 43.0\%
		& 0.487
		& 2.725 \\
		\bottomrule
	\end{tabular}
\caption{Routing efficiency on the complete 4,556-table pool using PaddleOCR-VL-1.6. Extra parser calls exclude initial parsing; controller calls exclude VC-Gate and VC-Ranker.}
	\label{tab:vc-gate-ablation}
\end{table*}
\subsection{Main Results}
As shown in Table~\ref{tab:main}, DEC consistently improves all three frozen base parsers under both controller configurations, with gains observed across all six source subsets. Qwen3.5-397B yields average improvements of 1.57 TEDS and 1.32 TEDS-S points across the three parsers, while the smaller Qwen3.5-35B still achieves average gains of 1.18 and 1.11 points, respectively. The larger controller generally provides stronger improvements, although the comparable results on MinerU2.5-Pro indicate that DEC does not depend on the largest controller. In contrast, direct parsing with either general VLM performs substantially worse than the specialized parsers on Real5 and Wild. These results show that DEC transfers across base parsers, data sources, and controller scales, whereas directly scaling a general VLM does not reliably address difficult real-world table parsing.

\subsection{Fine-Grained Analysis on TableParseMap}

Table~\ref{tab:tableparsemap-finegrained} compares eight direct-parsing baselines and three DEC-enhanced parsers across two aggregate error groups and five scenario categories, with results for the nine individual error types provided in Appendix~\ref{app:tableparsemap_results}. The direct baselines exhibit distinct capability profiles: MinerU2.5-Pro performs best overall and on the aggregated text and structural error groups, Qwen3.5-397B is strongest on Implicit Semantic and Large-Scale Tables, and Youtu achieves the strongest direct-parsing result on High-Homogeneity Tables. These variations show that aggregate performance alone obscures parser-specific strengths and weaknesses.

DEC improves the overall TEDS of MinerU2.5-Pro, PaddleOCR-VL-1.6, and GLM-OCR by 1.16, 1.90, and 2.61 points, respectively. The gains are consistently larger on structural errors than on text errors, indicating that DEC is particularly effective at repairing structural failures. Improvements are also pronounced on Large-Scale Tables, where all three parsers gain approximately 5--6 TEDS points, supporting the effectiveness of Decompose. The slight 0.24-point regression of MinerU2.5-Pro on text errors suggests that DEC does not uniformly improve every error category.
\subsection{Ablation Study}

\paragraph{Stage-wise Contribution.}
Table~\ref{tab:mechanism-ablation}(a) shows that Decompose, Enhance, and Correct provide successive improvements, with Correct contributing the largest incremental gain. Full DEC improves the parser baseline by 1.907 TEDS and 1.680 TEDS-S points. Although Decompose is triggered for only 178 of the 1,977 samples, it improves TEDS/TEDS-S by 3.202/2.575 points on this targeted subset, confirming its effectiveness on large tables despite its diluted full-set gain.

\paragraph{Visual-Consistency Ranking.}
Table~\ref{tab:mechanism-ablation}(b) shows that unconditionally accepting all proposals degrades both metrics after Enhance and results in a 29.34\% final regression rate. Agent Self-Judge filters many harmful updates but remains inferior to the independently trained VC-Ranker. VC-Ranker with rollback achieves the highest intermediate and final scores while reducing the regression rate to 12.65\%, demonstrating the importance of independent candidate verification.
\noindent

Additional controlled ablations show that fixed splitting and indiscriminate enhancement both degrade performance, confirming that the gains of Decompose and Enhance arise from agent-guided intervention rather than operation alone; detailed results are provided in Appendix~\ref{app:agent_guided_control}.
\subsection{Routing and Efficiency}

Table~\ref{tab:vc-gate-ablation} shows that VC-Gate triggers DEC on only 43.0\% of inputs while retaining 97.5\% of the TEDS gain and 97.8\% of the TEDS-S gain achieved by Always-on DEC. Compared with always-on execution, routing reduces extra parser calls by 47.3\% and controller calls by 34.9\%. Moreover, the routed subset improves from 71.454 to 74.763 TEDS, indicating that VC-Gate concentrates intervention on inputs that are substantially more likely to benefit from DEC. Detailed stage-wise iteration statistics further show that Enhance and Correct usually terminate well before their maximum budgets; see Appendix~\ref{app:iteration_statistics}.

\section{Conclusion}

We introduce TableParseMap, a diagnostic benchmark of 916 real-world
tables organized into five challenging scenarios and nine failure types.
Its fine-grained analysis reveals recurring limitations in processing
scale, visual evidence, and output consistency that motivate DEC, a
visual-consistency-guided agentic framework for frozen table parsers.
DEC uses a general VLM controller to progressively Decompose, Enhance,
and Correct parser outputs, while VC-Gate and VC-Ranker enable selective
intervention and verified updates without ground-truth HTML at inference
time. We further derive a 1,977-table Consensus-Hard Set for cross-model
hard-case evaluation. Across three frozen parsers and two controller
scales, DEC consistently improves parsing quality, achieving an average
gain of 1.57 TEDS points. These results support inference-time
collaboration between specialized parsers and general VLMs without
retraining. Limitations and future directions are discussed in
Appendix~\ref{app:limitations}.

\bibliographystyle{unsrtnat}
\bibliography{paper-references}
\clearpage
\appendix
\setcounter{secnumdepth}{1}
\section{TableParseMap: Construction and Fine-Grained Analysis}
\label{app:tableparsemap}

This section supplements the main paper with detailed descriptions of the benchmark construction and annotation protocol, complete category-wise results, and corresponding examples covering all five challenging scenarios and nine parser error types.

\subsection{Benchmark Construction and Annotation}
\label{app:tableparsemap_construction}

\paragraph{Scope and purpose.}
TableParseMap is a diagnostic benchmark containing 916
manually selected and annotated real-world table images. It is designed
to reveal both where table parsers struggle and how they fail.

At the benchmark level, TableParseMap provides two complementary
diagnostic views: five challenging scenarios and nine
representative parser failure types. To avoid overlapping evaluation
subsets and cross-category contamination, each image is assigned exactly
one canonical label according to its dominant difficulty factor. The 14
category-level subsets are therefore mutually exclusive.

\paragraph{Construction.}
We construct TableParseMap through a systematic review of errors produced
by multiple strong table parsers, including MinerU2.5-Pro,
Qwen3.5-397B-A17B, PaddleOCR-VL-1.6, and GLM-OCR. Their outputs are
manually compared with the corresponding table images to identify
recurring visual conditions and failure patterns. Multi-model outputs are
used to discover failure-prone samples and consolidate recurring
phenomena. The final samples
and labels are manually selected according to the dominant factor.

\paragraph{Ground-truth annotation and quality control.}
Ground-truth HTML is manually annotated from the source table images.
Annotators recover the row--column organization, header hierarchy,
merged-cell relations, and cell content. Each annotation is then manually
reviewed against the original image by another annotator. Disagreements or
annotation errors are corrected during review, while incomplete, duplicated,
or visually ambiguous samples are removed. Before evaluation, the final
annotations are normalized into a consistent HTML representation. Sensitive
information is masked during curation, and the redacted samples are manually
checked to ensure that masking does not remove content or structural evidence
required for evaluation.

\subsubsection{Where Parsers Struggle: Scenarios}
\label{app:tableparsemap_scenarios}

The five scenario subsets characterize complementary challenges arising
from visual quality, implicit structural cues, table scale, layout
irregularity, and local homogeneity. Their definitions and distributions
are summarized in Table~\ref{tab:tableparsemap_scenarios}.

\begin{table*}[t]
	\centering
	\small
	\setlength{\tabcolsep}{4pt}
	\begin{tabularx}{\textwidth}{@{}l r X@{}}
		\toprule
		Scenario & Samples & Definition \\
		\midrule
		Degraded Tables
		& 66
		& Table images are affected by quality degradation such as blur, low
		resolution, compression artifacts, background interference, or geometric
		distortion. These factors weaken the visibility of text, boundaries, and
		alignment cues, increasing the risk of content-recognition errors,
		structural reconstruction failures, and omissions. \\
		
		Implicit Semantic Tables
		& 52
		& Tables lack complete and explicit borders, and their row--column
		organization must be inferred not only from textual semantics but also
		from whitespace, text alignment, partial separators, and the plausibility
		of the overall layout. Local visual cues may be misleading, requiring
		global structural and semantic reasoning to determine the correct cell
		correspondences. \\
		
		Large-Scale Tables
		& 51
		& Tables have large spatial dimensions, many rows or columns, or highly
		dense content. Their scale may exceed the model's limited input resolution,
		visual-token budget, or effective perceptual range, resulting in omitted
		content, row--column misalignment, or incomplete structural reconstruction. \\
		
		Irregular Tables
		& 52
		& Tables deviate from conventional grid structures and may contain complex
		headers, asymmetric layouts, nested relations, local subtables, or unusual
		merged-cell patterns. The absence of stable structural regularities makes
		their row--column organization and hierarchy difficult to recover. \\
		
		High-Homogeneity Tables
		& 57
		& Neighboring rows, columns, cell contents, or blank regions exhibit highly
		similar and repetitive visual or textual patterns. Such homogeneity makes
		counting and correspondence difficult, frequently leading to misalignment,
		omission, or repeated output. \\
		\midrule
		\textbf{Total} & \textbf{278} & \\
		\bottomrule
	\end{tabularx}
	\caption{The five mutually exclusive scenario subsets in
		TableParseMap.}
	\label{tab:tableparsemap_scenarios}
\end{table*}

\subsubsection{How Parsers Fail: Error Types}
\label{app:tableparsemap_errors}

The nine failure types are divided into structural and textual errors.
Structural errors affect table topology, cell organization, or
row--column correspondence and therefore often have a substantial impact
on downstream usability. Textual errors are assigned only when the table
structure is essentially correct and the primary failure lies in
cell-content recognition.

\begin{table*}[t]
	\centering
	\small
	\setlength{\tabcolsep}{4pt}
	\begin{tabularx}{\textwidth}{@{}llrX@{}}
		\toprule
		Group & Error Type & Samples & Definition \\
		\midrule
		
		\multirow{5}{*}{\shortstack[l]{Structural\\Errors}}
		& Misalignment
		& 54
		& The table content is largely preserved without clear omission or
		redundancy, but is assigned to an incorrect row, column, or cell
		position. \\
		
		& Omission
		& 52
		& One or more cells, rows, columns, or local table regions visible
		in the source image are missing from the parsed output. \\
		
		& Redundancy
		& 50
		& The parsed output introduces additional cells, rows, columns,
		structural regions, or content that are not supported by the source
		image. \\
		
		& Repetition
		& 46
		& The same source cell, row, column, or table fragment is generated
		more than once. Although repetition can be regarded as a specific
		form of redundancy, it is treated separately because of its
		distinctive appearance and diverse triggering conditions. \\
		
		& Merged-cell Error
		& 105
		& The parser reconstructs \texttt{rowspan}, \texttt{colspan}, or
		other merged-cell relations incorrectly, including under-merging,
		where cells that should be merged remain separated, and
		over-merging, where distinct cells are incorrectly merged. \\
		
		\cmidrule(lr){2-4}
		
		\multirow{4}{*}{\shortstack[l]{Textual\\Errors}}
		& Formula Error
		& 154
		& Mathematical formulas, their internal structures, subscripts,
		superscripts, operators, or formula symbols are recognized
		incorrectly. \\
		
		& Special-symbol Error
		& 87
		& Special symbols, dashed or underlined content, list markers,
		units, or uncommon characters are omitted, substituted, or
		confused, such as confusion between \texttt{<li>} and punctuation. \\
		
		& Plain-text Error
		& 36
		& Ordinary text, numbers, or punctuation are recognized incorrectly
		while the table structure and cell correspondence remain correct.
		This category includes errors involving handwriting, uncommon
		characters, character deformation, and other general OCR failures. \\
		
		& Line-break Error
		& 54
		& Line breaks, paragraph boundaries, or multi-line organization
		within a cell are reconstructed incorrectly, such as splitting text
		that should remain continuous or merging numbered items that should
		remain on separate lines. \\
		
		\midrule
		\multicolumn{2}{l}{\textbf{Total}}
		& \textbf{638}
		& \\
		\bottomrule
	\end{tabularx}
	
	\caption{The nine mutually exclusive parser failure types in
		TableParseMap. Structural errors affect table topology or the
		correspondence between content and cells, whereas textual errors
		occur when the table structure is essentially correct but the
		content within cells is recognized or organized incorrectly.
		The benchmark contains 307 structural-error samples and 331
		textual-error samples.}
	\label{tab:tableparsemap_errors}
\end{table*}

\paragraph{Structural-first annotation rule.}
When structural and textual symptoms co-occur, TableParseMap follows a
structural-first rule. Any failure that changes row, column, cell,
alignment, omission, repetition, or span relations is assigned to the
structural group, even when text is consequently missing or misplaced.
A textual label is used only when the table structure remains correct
and the dominant error lies in the recognized cell content. When multiple
difficulty factors are present, annotators select the factor that most
directly explains the dominant parsing difficulty. This single-label
design prevents the same sample from being counted repeatedly across
correlated categories.

Upon publication, we will release the TableParseMap annotations,
category definitions, benchmark manifests, and evaluation code.

\subsection{Complete Fine-Grained Results}
\label{app:tableparsemap_results}

Table~\ref{tab:tableparsemap-finegrained} reports results on all 916
TableParseMap samples, including the two aggregate output-error groups
and all five challenging input scenarios. We further decompose the
structural- and textual-error groups into five structural error types
and four textual error types in
Tables~\ref{tab:tableparsemap-structural-errors}
and~\ref{tab:tableparsemap-textual-errors}, respectively.
All $+$DEC rows use Qwen3.5-397B-A17B as the controller with
VC-Gate routing.

The aggregate results reveal distinct parser-specific capability
profiles. MinerU2.5-Pro performs best among direct parsers overall and
on both aggregate error groups, while Qwen3.5-397B performs strongly on
Implicit Semantic and Large-Scale Tables. DEC improves all three frozen
parsers overall, with its gains concentrated primarily on structural
errors and difficult large-scale inputs.

The error-type results provide a more detailed view of these
improvements. As shown in
Table~\ref{tab:tableparsemap-structural-errors}, DEC substantially
improves repetition and merged-cell errors across the frozen parsers.
PaddleOCR-VL-1.6 improves on all five structural error types, while
GLM-OCR obtains particularly large gains on misalignment and repetition.
For MinerU2.5-Pro, the largest structural gains occur on repetition and
merged-cell errors.

In contrast, the textual-error results in
Table~\ref{tab:tableparsemap-textual-errors} are more mixed.
DEC improves most textual categories for PaddleOCR-VL-1.6 and GLM-OCR,
but slightly reduces MinerU2.5-Pro on formula, plain-text, and line-break
errors. This explains the small regression of MinerU2.5-Pro on the
aggregate textual-error group and indicates that inference-time
intervention is more consistently beneficial for structural recovery
than for already strong text recognition.
\begin{table*}[t]
	\centering
	\small
	\setlength{\tabcolsep}{1.35mm}
	\renewcommand{\arraystretch}{1.04}
	
	\begin{tabular}{@{}l*{5}{c}@{}}
		\toprule
		Parser
		& \shortstack{Misalignment\\($N=54$)}
		& \shortstack{Omission\\($N=52$)}
		& \shortstack{Redundancy\\($N=50$)}
		& \shortstack{Repetition\\($N=46$)}
		& \shortstack{Merged-cell\\Error ($N=105$)} \\
		\midrule
		
		MinerU2.5-Pro
		& 87.30/90.01
		& \textbf{87.10}/\underline{90.80}
		& \underline{89.91}/\underline{91.90}
		& \underline{84.69}/87.16
		& 80.39/84.73 \\
		
		PaddleOCR-VL-1.6
		& 85.52/89.79
		& 83.21/87.96
		& 85.80/89.43
		& 77.96/82.83
		& 79.58/84.20 \\
		
		GLM-OCR
		& 79.20/82.64
		& 83.85/88.47
		& 86.23/88.58
		& 77.28/81.31
		& 78.23/82.39 \\
		
		Qwen3.5-397B
		& 80.60/85.18
		& 83.19/87.17
		& 77.29/81.01
		& 35.45/36.04
		& 79.61/84.70 \\
		
		Qwen3.5-35B
		& 86.91/\underline{90.46}
		& 82.37/86.18
		& 87.68/91.60
		& 83.20/85.18
		& 80.31/85.19 \\
		
		DeepSeek-OCR-2
		& 57.93/66.38
		& 64.23/71.98
		& 58.50/64.39
		& 33.37/38.09
		& 59.92/68.40 \\
		
		Unlimited-OCR
		& 72.81/77.73
		& 73.37/77.61
		& 70.97/75.89
		& 57.26/63.19
		& 63.16/68.09 \\
		
		Youtu-Parsing
		& 83.62/87.71
		& 81.31/85.57
		& 83.91/87.61
		& 66.32/72.21
		& 78.39/83.97 \\
		
		\midrule
		
		MinerU2.5-Pro $+$ DEC
		& \underline{87.50}/90.08
		& \underline{87.07}/\textbf{90.99}
		& \textbf{90.92}/\textbf{92.57}
		& \textbf{87.38}/\textbf{89.66}
		& \textbf{83.69}/\textbf{88.22} \\
		
		PaddleOCR-VL-1.6 $+$ DEC
		& \textbf{87.83}/\textbf{91.63}
		& 84.55/89.48
		& 89.25/91.63
		& 81.72/86.02
		& \underline{82.17}/\underline{86.78} \\
		
		GLM-OCR $+$ DEC
		& 85.57/88.81
		& 84.64/89.04
		& 86.93/88.79
		& 82.56/\underline{87.17}
		& 82.05/86.13 \\
		
		\bottomrule
	\end{tabular}
	
	\caption{Complete TEDS/TEDS-S results on the five structural-error
		types in TableParseMap. $+$DEC uses Qwen3.5-397B with VC-Gate.
		Best and second-best values are bold and underlined, respectively.}
	\label{tab:tableparsemap-structural-errors}
\end{table*}
\begin{table*}[t]
	\centering
	\small
	\setlength{\tabcolsep}{2.3mm}
	\renewcommand{\arraystretch}{1.04}
	
	\begin{tabular}{@{}l*{4}{c}@{}}
		\toprule
		Parser
		& \shortstack{Formula\\Error ($N=154$)}
		& \shortstack{Special-symbol\\Error ($N=87$)}
		& \shortstack{Plain-text\\Error ($N=36$)}
		& \shortstack{Line-break\\Error ($N=54$)} \\
		\midrule
		
		MinerU2.5-Pro
		& \textbf{85.32}/\textbf{94.76}
		& \underline{84.32}/\underline{89.46}
		& \textbf{86.13}/\textbf{94.22}
		& \textbf{91.37}/\textbf{93.12} \\
		
		PaddleOCR-VL-1.6
		& 85.12/93.34
		& 79.15/85.58
		& 80.66/89.29
		& 85.26/\underline{87.63} \\
		
		GLM-OCR
		& 81.57/91.13
		& 79.64/85.82
		& 80.36/90.10
		& 85.73/87.51 \\
		
		Qwen3.5-397B
		& 80.10/90.81
		& 77.52/83.74
		& 76.41/84.11
		& 84.44/86.22 \\
		
		Qwen3.5-35B
		& 80.62/89.77
		& 75.09/80.98
		& 82.22/87.66
		& 78.45/81.63 \\
		
		DeepSeek-OCR-2
		& 63.83/75.60
		& 55.57/62.00
		& 59.38/72.18
		& 61.74/66.97 \\
		
		Unlimited-OCR
		& 67.00/77.85
		& 48.56/53.20
		& 66.70/73.24
		& 71.51/73.73 \\
		
		Youtu-Parsing
		& 80.02/91.69
		& 71.67/77.98
		& 72.37/82.47
		& 79.20/82.02 \\
		
		\midrule
		
		MinerU2.5-Pro $+$ DEC
		& 84.98/\underline{94.57}
		& \textbf{84.59}/\textbf{89.71}
		& \underline{84.87}/\underline{92.83}
		& \underline{91.31}/\textbf{93.12} \\
		
		PaddleOCR-VL-1.6 $+$ DEC
		& \underline{85.13}/93.82
		& 80.55/86.59
		& 81.42/89.57
		& 85.36/\underline{87.63} \\
		
		GLM-OCR $+$ DEC
		& 83.44/93.24
		& 79.75/86.43
		& 81.31/90.47
		& 85.29/87.25 \\
		
		\bottomrule
	\end{tabular}
	
	\caption{Complete TEDS/TEDS-S results on the four textual-error
		types in TableParseMap. $+$DEC uses Qwen3.5-397B with VC-Gate.
		Best and second-best values are bold and underlined, respectively.}
	\label{tab:tableparsemap-textual-errors}
\end{table*}

\subsection{Representative Cases}
\label{app:tableparsemap_cases}

\begin{figure*}[p]
	\centering
	\includegraphics[
	height=0.84\textheight,
	keepaspectratio
	]{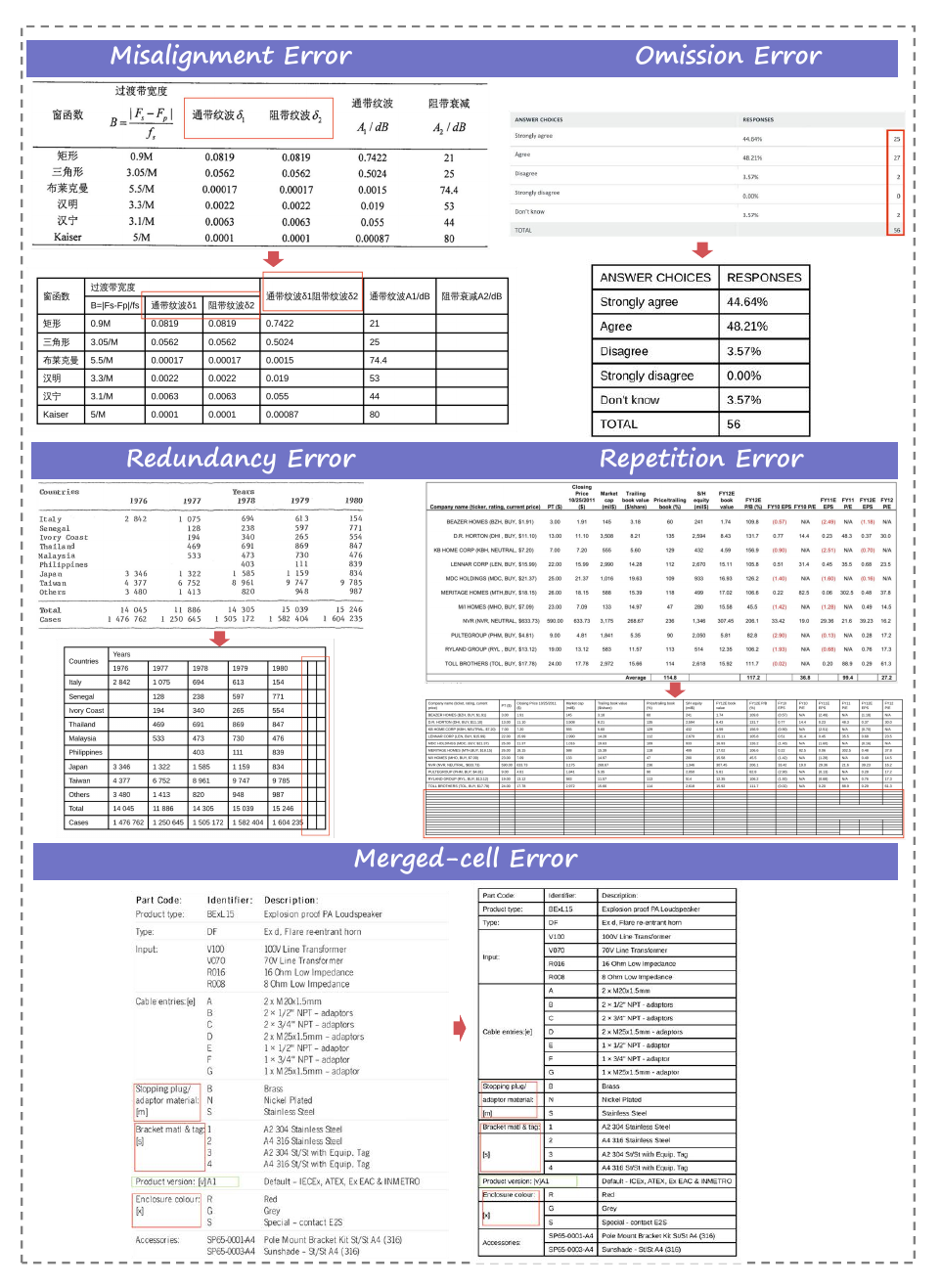}
	\caption{
		Representative cases of the five structural-error types in
		TableParseMap. Misalignment preserves most table content but assigns
		it to incorrect rows, columns, or cells. Omission removes visible
		cells or structural regions, whereas Redundancy introduces content or
		structure unsupported by the source image. Repetition emits the same
		source region more than once. Merged-cell Error incorrectly recovers
		cell-span relations through under-merging or over-merging. Highlighted
		regions mark the dominant structural discrepancy in each example.
	}
	\label{fig:tableparsemap_structural_cases}
\end{figure*}

\begin{figure*}[p]
	\centering
	\includegraphics[
	height=0.84\textheight,
	keepaspectratio
	]{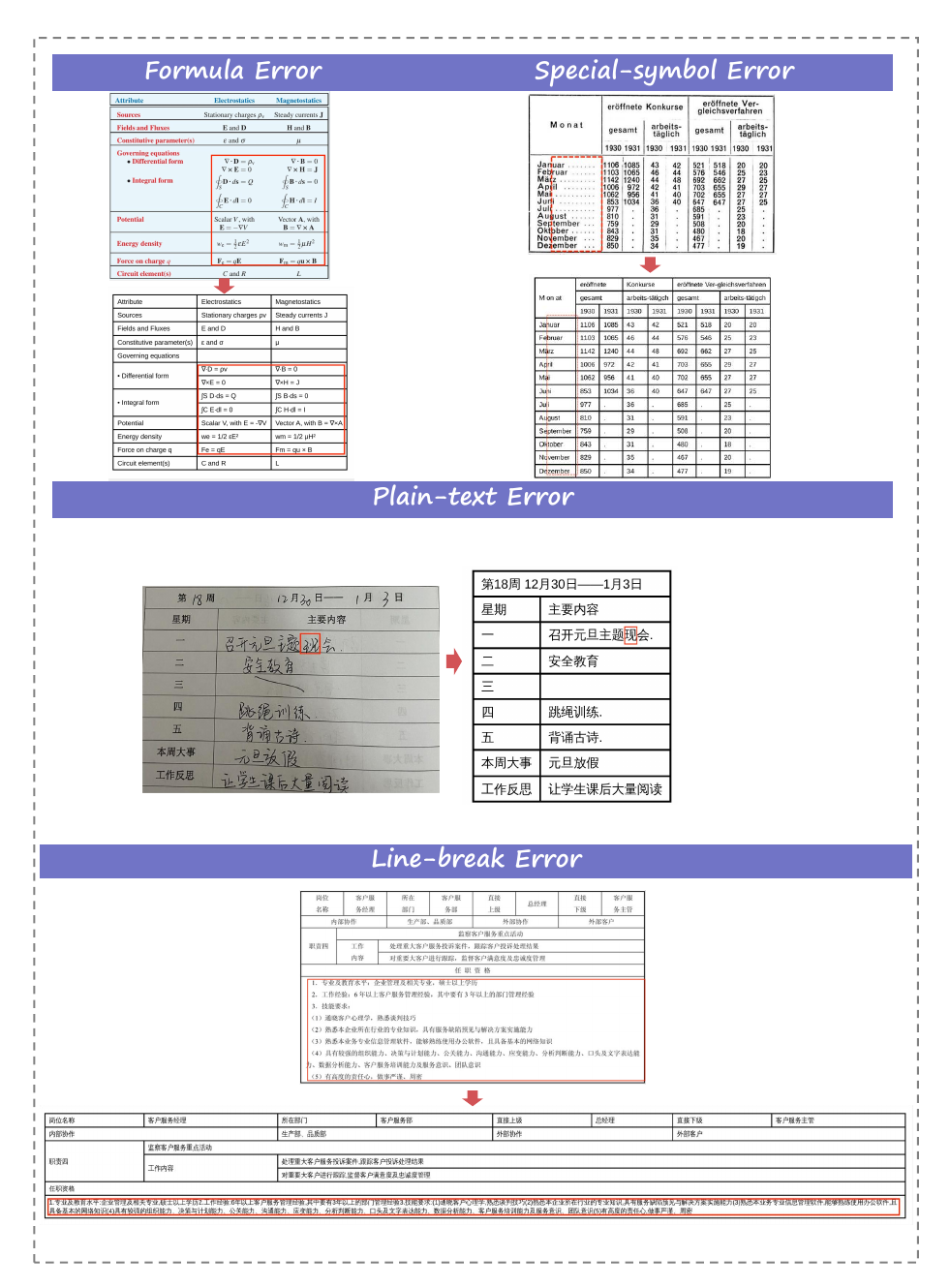}
	\caption{
		Representative cases of the four textual-error types in
		TableParseMap. Formula Error covers incorrect transcription of
		mathematical expressions or their internal structure.
		Special-symbol Error includes omitted or confused symbols, markers,
		underlines, and other uncommon characters. Plain-text Error captures
		ordinary OCR failures involving text, numbers, punctuation, or
		handwritten content. Line-break Error concerns incorrect organization
		of multi-line content within an otherwise structurally correct cell.
		Highlighted regions indicate the corresponding content-recognition
		error.
	}
	\label{fig:tableparsemap_textual_cases}
\end{figure*}

Figures~\ref{fig:tableparsemap_structural_cases}
and~\ref{fig:tableparsemap_textual_cases} provide representative
examples of all nine parser failure types in TableParseMap. Each case
contrasts the source table with a parser output, while the highlighted
regions indicate the dominant discrepancy used for category assignment.

As illustrated in Figure~\ref{fig:tableparsemap_textual_cases}, textual
categories are assigned only when the table structure and cell
correspondence remain essentially correct. The distinction therefore
depends on the affected content representation: formulas preserve their
special internal syntax, special-symbol errors concern non-standard
visual tokens, plain-text errors cover general character recognition,
and line-break errors alter the organization of otherwise recognized
multi-line content.

\section{Training and Evaluation Details}
\label{app:training_details}

\subsection{Evaluation Protocol and Metrics}
\label{app:evaluation_details}

\paragraph{Evaluation settings.}
We use three evaluation sets for complementary purposes. The main comparison is
conducted on the 1,977-table Consensus-Hard Set, which focuses on samples that
remain difficult for multiple heterogeneous parsers. Fine-grained behavior analysis
is performed on the complete 916-table TableParseMap. Routing and efficiency are
evaluated on the complete six-source evaluation pool of 4,556 tables, using
PaddleOCR-VL-1.6 as the base parser.

The evaluation pool contains six sources: CCOCR, the In-house TR Benchmark,
OCRBenchv2, TableParseMap, Real5, and Wild.
CCOCR and OCRBenchv2 are established public benchmarks. Real5 and Wild
originate from page-level OmniDocBench data rather than isolated table
images; MinerU and PP-DocLayoutV3 are used
to detect table regions, and GPT-5.2 assists in verifying crop
completeness and crop--HTML correspondence.
\paragraph{DEC evaluation configuration.}
We evaluate DEC with MinerU2.5-Pro, PaddleOCR-VL-1.6, and GLM-OCR as
frozen base parsers. Unless otherwise stated, Qwen3.5-397B-A17B is used
as the frozen controller. The controller adopts deterministic decoding
with reasoning mode disabled. For each table or
decomposed region, Enhance invokes four evidence-preserving
tools, while Correct performs at most three repair rounds. A candidate
update is accepted only when its VC-Ranker score exceeds that of the
current result by a margin of
\(\delta=0.05\); otherwise, the previous result is retained. The same
margin is used to verify merged outputs when Decompose is activated.
\paragraph{Consensus-Hard construction.}
We run MinerU2.5-Pro, PaddleOCR-VL-1.6, GLM-OCR, and
Qwen3.5-397B-A17B on every candidate in the 4,556-table source pool. For
sample $I$, let $m_j(I)$ denote the TEDS score obtained by the $j$-th
parser. A sample is selected when at least two parsers score below 0.90:

\begin{equation}
	h(I)
	=
	\mathbb{I}
	\left[
	\sum_{j=1}^{4}
	\mathbb{I}\left(m_j(I)<0.90\right)
	\geq 2
	\right]
	\label{eq:consensus_hard_selection}
\end{equation}

Requiring failure from multiple heterogeneous parsers reduces the
influence of an isolated model-specific weakness. Table~\ref{tab:consensus_composition} reports the final source
composition.

\begin{table}[t]
	\centering
	\small
	\setlength{\tabcolsep}{5pt}
	\begin{tabular}{@{}lrr@{}}
		\toprule
		Source & Source Pool & Selected \\
		\midrule
		CCOCR & 300 & 118 \\
		OCRBenchv2 & 694 & 259 \\
		In-house TR & 413 & 278 \\
		TableParseMap & 916 & 602 \\
		Real5 & 1,902 & 614 \\
		Wild & 331 & 106 \\
		\midrule
		\textbf{Total} & \textbf{4,556} & \textbf{1,977} \\
		\bottomrule
	\end{tabular}
	\caption{Source composition of the Consensus-Hard Set.}
	\label{tab:consensus_composition}
\end{table}

\paragraph{TEDS and TEDS-S.}
TEDS measures similarity between the tree representations of the
predicted and ground-truth HTML tables \cite{smock2022pubtables}. Let
$T_{\mathrm{pred}}$ and $T_{\mathrm{gt}}$ denote the corresponding trees
and $d_{\mathrm{tree}}(\cdot,\cdot)$ their tree-edit distance. We compute

\begin{equation}
	\operatorname{TEDS}
	=
	1-
	\frac{
		d_{\mathrm{tree}}
		\left(T_{\mathrm{pred}},T_{\mathrm{gt}}\right)
	}{
		\max\left(
		\lvert T_{\mathrm{pred}}\rvert,
		\lvert T_{\mathrm{gt}}\rvert
		\right)
	}
	\label{eq:teds_appendix}
\end{equation}

TEDS jointly reflects table structure and cell content. TEDS-S uses the
same normalized tree-edit formulation but removes cell-text differences,
thereby focusing on row--column organization, node presence, and
\texttt{rowspan}/\texttt{colspan} relations. Both metrics range from 0
to 1 and are multiplied by 100 in all reported tables.

Before scoring, predicted and ground-truth HTML are normalized into the
same canonical representation. For a dataset $\mathcal{D}$ containing
$N$ samples, the reported score is

\begin{equation}
	M(\mathcal{D})
	=
	\frac{100}{N}
	\sum_{i=1}^{N} M_i,
	\quad
	M\in\left\{
	\operatorname{TEDS},
	\operatorname{TEDS\text{-}S}
	\right\}
	\label{eq:dataset_metric}
\end{equation}

For Consensus-Hard, Overall is the sample-weighted average over all
1,977 samples, so each table contributes equally regardless of its
source subset.

\paragraph{Efficiency accounting.}
The \emph{DEC trigger rate} is the proportion of inputs routed beyond the
initial parser result. \emph{Extra parser calls per input} exclude the
initial whole-image parsing call shared by the parser-only baseline and
DEC, and count additional calls caused by decomposition or reparsing
transformed views. \emph{Controller calls per input} count actual
invocations of the general VLM during Decompose, Enhance, and Correct. Stage-wise invocation statistics report the
actual number of Enhance tool calls and Correct diagnosis/repair rounds,
rather than their maximum permitted budgets.

\subsection{Training of VC-Gate and VC-Ranker}
\label{app:vc_training}

Both VC-Gate and VC-Ranker are initialized from
Qwen3-VL-8B-Instruct and trained through full-parameter optimization.
Both models are trained for three epochs on eight GPUs with a
per-device batch size of one. We use AdamW with a base learning rate of
$1\times10^{-5}$, a vision-encoder learning rate of $1\times10^{-6}$,
a weight decay of 0.01, and cosine decay. Training uses
BF16 precision and DeepSpeed ZeRO-3.

VC-Gate is trained as a binary visual-consistency classifier on
approximately 44K image--render examples. Candidates with offline TEDS
at least 0.95 are treated as positive examples, candidates below 0.85 are
treated as negative examples, and candidates in the ambiguous interval
$[0.85,0.95)$ are excluded. VC-Ranker is trained as a scalar reward model
on approximately 21K pairwise comparisons using a Bradley--Terry ranking
objective. Candidate pairs are constructed from different predictions of
the same reference table, and only pairs whose TEDS scores differ by at
least 0.20 are retained. The higher-scoring candidate is treated as the
preferred result.

For both corpora, 1\% of the data is held out for validation. All
candidates derived from the same reference image are assigned to the
same split, preventing reference-image overlap between training and
validation.

\section{Agent-Guided Control}
\label{app:agent_guided_control}

As shown in Table~\ref{tab:mechanism-ablation}(a), both Decompose and
Enhance provide incremental gains to the complete framework. We further
examine whether these gains arise from informed agent decisions or
merely from applying splitting and enhancement operations.

\begin{table*}[t]
	\centering
	\small
	\renewcommand{\arraystretch}{1.05}
	\begin{minipage}[t]{0.48\textwidth}
		\centering
		\textbf{(a) Decomposition Strategy}\\[2pt]
		\setlength{\tabcolsep}{3.5pt}
		\resizebox{\linewidth}{!}{%
		\begin{tabular}{@{}lcc@{}}
			\toprule
			Strategy
			& TEDS / TEDS-S
			& \shortstack{Gain over No Split\\TEDS / TEDS-S} \\
			\midrule
			No Split
			& 82.565 / 88.841
			& -- \\
			Fixed Split
			& 81.960 / 87.569
			& $-0.605$ / $-1.272$ \\
			Agent-Guided Split
			& \textbf{87.268 / 92.167}
			& \textbf{$+4.703$ / $+3.326$} \\
			\bottomrule
		\end{tabular}%
		}
	\end{minipage}
	\hfill
	\begin{minipage}[t]{0.48\textwidth}
		\centering
		\textbf{(b) Enhancement Strategy}\\[2pt]
		\setlength{\tabcolsep}{3.5pt}
		\resizebox{\linewidth}{!}{%
		\begin{tabular}{@{}lcc@{}}
			\toprule
			Strategy
			& TEDS / TEDS-S
			& \shortstack{Gain over No Enhance\\TEDS / TEDS-S} \\
			\midrule
			No Enhance
			& 67.791 / 78.128
			& -- \\
			Agent Selective
			& \textbf{68.323 / 78.548}
			& \textbf{$+0.532$ / $+0.420$} \\
			Always Apply All
			& 65.980 / 76.538
			& $-1.811$ / $-1.590$ \\
			\bottomrule
		\end{tabular}%
		}
	\end{minipage}
	\caption{Ablations of agent-guided control. (a) Decomposition
		strategies on the 51-table Large-Scale subset. (b) Enhancement
		strategies on the Consensus-Hard Set. All experiments use
		PaddleOCR-VL-1.6 with Qwen3.5-397B-A17B.}
	\label{tab:agent-guided-control}
\end{table*}

\paragraph{Agent-guided decomposition.}
Table~\ref{tab:agent-guided-control}(a) examines whether the benefit of
Decompose comes from splitting itself or from selecting safe structural
boundaries. On the same 51 large tables, fixed splitting decreases TEDS
and TEDS-S by 0.605 and 1.272 points, respectively, indicating that
mechanical partitioning can cut through local content or disrupt
cross-region structure. In contrast, agent-guided splitting improves the
unsplit baseline by 4.703 TEDS and 3.326 TEDS-S points and outperforms
fixed splitting by 5.307 and 4.598 points. Effective decomposition
therefore depends on visual reasoning over the table structure rather
than merely reducing the image size.

\paragraph{Agent-guided enhancement.}
Table~\ref{tab:agent-guided-control}(b) evaluates whether Enhance
benefits from selective tool use rather than image transformation alone.
Starting from the same Stage-2 inputs, Agent Selective improves No
Enhance by 0.532 TEDS and 0.420 TEDS-S points. Applying all enhancement
tools indiscriminately instead reduces the scores by 1.811 and 1.590
points, causing Agent Selective to outperform it by 2.343 TEDS and 2.010
TEDS-S points. Visual transformations are therefore not universally
beneficial; the controller must match each tool to the available visual
evidence and avoid unnecessary intervention.

\section{Routing and Efficiency Analysis}
\label{app:routing_efficiency}

\subsection{VC-Gate Routing}
\label{app:vc_gate_routing}

Table~\ref{tab:vc-gate-ablation} compares parser-only execution,
Always-on DEC, and VC-Gate routing on the complete 4,556-table pool.

VC-Gate triggers DEC on only 43.0\% of inputs while retaining 97.5\% of
the TEDS gain and 97.8\% of the TEDS-S gain achieved by Always-on DEC.
Relative to always-on execution, routing reduces extra parser calls by
47.3\% and controller calls by 34.9\%. The remaining differences of only
0.036 TEDS and 0.027 TEDS-S show that most reliable first-pass outputs
can bypass further intervention without sacrificing the overall benefit
of DEC.

\subsection{Stage-Wise Invocation Statistics}
\label{app:iteration_statistics}

Beyond input-level routing, we measure how frequently the controller
terminates Enhance and Correct before reaching their maximum budgets.

\begin{figure*}[t]
	\centering
	\includegraphics[width=\textwidth]
	{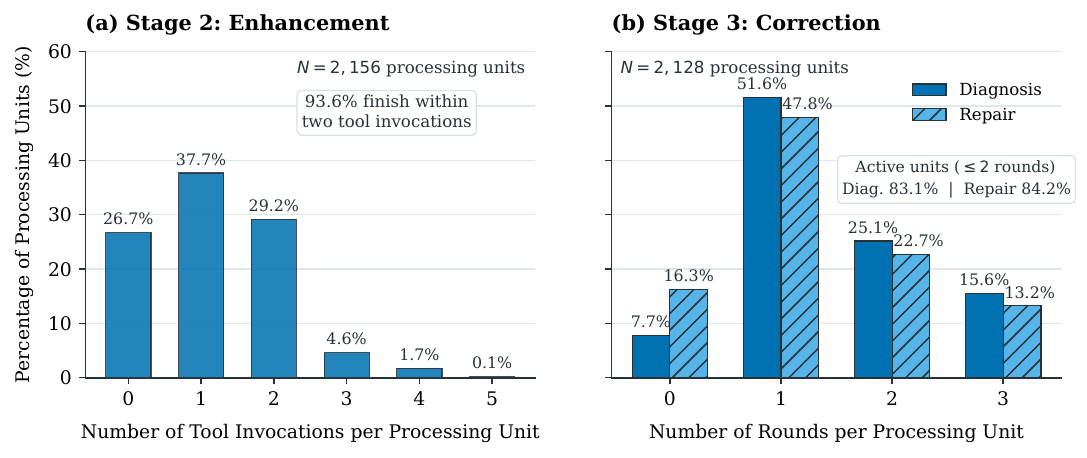}
	\caption{Distribution of the actual computation executed by Enhance
		and Correct. A processing unit is either an unsplit table or a region
		produced by Decompose. Left: enhancement-tool invocations. Right:
		diagnosis and repair rounds during Correct. Percentages above the
		bars are computed over all processing units; the right-panel
		annotation reports proportions among active units.}
	\label{fig:dec_iteration_distribution}
\end{figure*}

As shown in Figure~\ref{fig:dec_iteration_distribution}, 26.7\% of the
2,156 Enhance processing units require no image transformation, while
37.7\% and 29.2\% terminate after one and two tool invocations,
respectively. Thus, 93.6\% finish within at most two invocations, only
6.4\% require more than two, and 0.1\% reach the maximum budget of five.
The controller therefore usually identifies a useful transformation or
terminates the stage without exhausting the available tools.

Correct exhibits a similar early-stopping pattern over 2,128 processing
units. One round is the most common outcome, accounting for 51.6\% of
diagnosis sequences and 47.8\% of repair sequences. Only 15.6\% and
13.2\% reach the maximum of three diagnosis and repair rounds,
respectively. Among units on which the corresponding operation is
activated, 83.1\% of diagnosis sequences and 84.2\% of repair sequences
finish within two rounds. Together with VC-Gate routing, these results
show that DEC concentrates computation on difficult inputs while usually
terminating well before its maximum inference budget.

\section{Limitations and Future Work}
\label{app:limitations}

DEC incurs additional parser and controller calls compared with direct
parsing, although VC-Gate substantially reduces unnecessary
intervention. Its effectiveness also depends on the reliability of
VC-Gate and VC-Ranker and on the fidelity of HTML rendering, which may
cause beneficial candidates to be rejected or harmful updates to be
accepted. Moreover, TEDS-based metrics may over-penalize local structural
deviations that have limited impact on readability or downstream
utility. Future work will investigate lighter controllers, stronger
visual-consistency models, finer-grained tool-routing policies, and
evaluation protocols that jointly consider structural accuracy, visual
fidelity, and task utility.
\section{DEC Tool Implementation Details}
\label{app:tool_implementation}

DEC uses one decomposition tool in Stage~1 and four visual-intervention
tools in Stage~2. Table~\ref{tab:dec_enabled_tools} summarizes the tools in experiments. Stage~3 does not invoke an external
tool; instead, the controller directly diagnoses and edits the current
structured prediction.

\begin{table*}[t]
	\centering
	\small
	\setlength{\tabcolsep}{4pt}
	\begin{tabularx}{\textwidth}{@{}llXX@{}}
		\toprule
		Stage & Tool & Purpose & Main implementation \\
		\midrule
		
		Decompose
		& \texttt{split\_table}
		& Partitions large tables into locally tractable regions.
		& Estimates table scale, lets the controller inspect and adjust
		candidate split boundaries. \\
		
		Enhance
		& \texttt{semantic\_scaffold}
		& Makes implicit semantic regions more explicit.
		& The controller proposes normalized regions corresponding to
		headers, section rows, stub columns, or metric groups. The regions
		are indicated by translucent overlays. \\
		
		Enhance
		& \texttt{table\_roi\_refiner}
		& Removes irrelevant context outside the table.
		& The controller predicts a table bounding box, optional padding,
		and exclusion regions. \\
		
		Enhance
		& \texttt{image\_quality\_normalizer}
		& Improves weak contrast and locally degraded visual evidence.
		& Applies image statistics, automatic contrast
		adjustment, and mild sharpening when the measured visual quality
		is insufficient; otherwise, it returns the input unchanged. \\
		
		Enhance
		& \texttt{cell\_anchor}
		& Disambiguates repetitive rows and visually similar cells.
		& Detects a grid, recognizes selected cells, temporarily
		replaces their contents with unique anchors, reparses the modified
		image, and restores the original contents in the resulting HTML. \\
		
		\bottomrule
	\end{tabularx}
	\caption{Tools enabled in theDEC experiments.}
	\label{tab:dec_enabled_tools}
\end{table*}

\paragraph{Structure-aware decomposition.}
The \texttt{split\_table} tool is considered when the estimated table
scale exceeds a predefined threshold. Rather than applying a fixed cut,
the controller inspects the table layout and adjusts candidate boundaries
to avoid cutting through rows, cells, or spanning structures. Each
resulting region independently passes through Enhance and Correct. The
final regional HTML predictions are converted to OTSL, vertically aligned,
and merged into a complete table. For decomposed inputs only, the merged
prediction is finally compared with the immutable whole-image baseline,
and the baseline is retained when the merge cannot be reliably verified.

\paragraph{Reference--working view separation.}
Stage~2 maintains an immutable reference image \(I^{\mathrm{ref}}\) and
a mutable working image \(I^{\mathrm{work}}\). At each round, the
controller either selects one enhancement tool or terminates the stage,
with at most five tool invocations per processing unit. After a tool
modifies \(I^{\mathrm{work}}\), the frozen base parser is immediately
rerun to produce a new HTML candidate. The current and proposed
predictions are rendered, and VC-Ranker compares both candidates against
the same immutable reference image. A proposal is committed only when

\begin{equation}
	s_{\mathrm{proposal}}
	>
	s_{\mathrm{current}}+\delta,
\end{equation}

where \(\delta\) follows the acceptance margin specified in the final
evaluation configuration. Otherwise, both the image and HTML are rolled
back to the previous accepted state. The controller receives only a
compact execution history containing the selected tool, its arguments,
a result summary, the consistency scores, and the acceptance decision.

\paragraph{Enhancement tools.}
The \texttt{semantic\_scaffold} tool introduces sparse visual cues for
semantically meaningful regions but does not insert textual labels or
directly edit HTML. The \texttt{table\_roi\_refiner} removes only
clearly irrelevant context outside the table; near-full-image crops are
treated as no-ops, while excessively small or incomplete regions are
rejected. The \texttt{image\_quality\_normalizer} is a
image-processing procedure. It
uses image statistics to determine whether automatic contrast adjustment
or mild sharpening is required.

The \texttt{cell\_anchor} tool targets tables
with repetitive or visually ambiguous cells. It constructs candidate cell regions, and temporarily
replaces a small number of reliably recognized cells with unique markers.
After reparsing, the markers are replaced with their original contents
in HTML. The complete proposal is rejected when an expected marker is
missing, duplicated, or cannot be restored unambiguously.

\end{document}